# Generating Context-Appropriate Word Orders in Turkish

Beryl Hoffman*
Department of Computer and Information Sciences
University of Pennsylvania
(hoffman@linc.cis.upenn.edu)

## 1 Introduction

Turkish, like Finnish, German, Hindi, Japanese, and Korean, has considerably freer word order than English. In these languages, word order variation is used to convey distinctions in meaning that are not generally captured in the semantic representations that have been developed for English, although these distinctions are also present in somewhat less obvious ways in English. In the next section, I present a summary of the linguistic data on Turkish word order variations. Section 3 describes the categorial formalism I propose to model the syntax, semantics, and pragmatic information in Turkish sentences. To capture the syntax of free word order languages, I present an adaptation of Combinatory Categorial Grammars, CCGs (Steedman-85; Steedman-91), called {}-CCGs (set-CCGs). Then, I integrate a level of information structure, representing pragmatic functions such as topic and focus, with {}-CCGs to allow pragmatic distinctions in meaning to influence the word order of the sentence in a compositional way. In Section 4, I discuss how this strategy is used within a generation system which produces Turkish sentences with word orders appropriate to the context, and include sample runs of the implementation.

## 2 Free Word Order in Turkish

The most common word order used in simple transitive sentences in Turkish is SOV (Subject-Object-Verb), but all six permutations of a transitive sentence can be used in the proper discourse situation since the subject and object are differentiated by case-marking.[1]

(1) a. Ayşe Fatma'yı arıyor.
    Ayşe Fatma-Acc seek-Pres-(3Sg).
    "Ayşe is looking for Fatma."
  b. Fatma'yı Ayşe arıyor.
  c. Ayşe arıyor Fatma'yı.
  d. Fatma'yı arıyor Ayşe.
  e. Arıyor Fatma'yı Ayşe.
  f. Arıyor Ayşe Fatma'yı.

The propositional interpretation assigned to all six of these sentences is *seek'(Ayşe',Fatma')*. However, each word order conveys a different discourse meaning only appropriate to a specific discourse situation. The one propositional interpretation cannot capture the distinctions in meaning necessary for effective translation and communication in Turkish. The interpretations of these different word orders rely on discourse-related notions such as *theme/rheme, focus/presupposition, topic/comment*, etc. that describe how the sentence relates to its context.

There is little agreement on how to represent the discourse-related functions of components in the sentence, i.e. the information structure of the sentence. Among Turkish linguists, Erguvanlı (Erguvanli-84) captures the general use of word order by associating each position in a Turkish sentence with a specific pragmatic function. Generally in Turkish, speakers first place the information that links the sentence to the previous context, then the important and/or new information immediately before the verb, and the information that is not really needed but may help the hearer understand the sentence better, after the verb. Erguvanlı identifies the sentence-initial position as the *topic*, the immediately preverbal position as the *focus*, and the postverbal positions as *backgrounded information*. The following template that I will be using in the implementation describes the general association between sentence positions and information structure components (in bold font) for Turkish:

(2) **Topic Neutral Focus** Verb **Background**

---

*I would like to thank Mark Steedman, Miriam Butt, and the anonymous referees for their valuable advice. This work was partially supported by DARPA N00014-90-J-1863, ARO DAAL03-89-C-0031, NSF IRI 90-16592, Ben Franklin 91S.3078C-1.

[1] According to a language acquisition study in (Slobin-82), 52% of transitive sentences used by a sample of Turkish speakers were not in the canonical SOV word order.

I will call the phrase formed by the *topic* and the *neutral* components the *theme* of the sentence and the phrase formed by the *focus* and the verb, the *rheme* of the sentence.

Using these information structure components, we can now explain why certain word orders are appropriate or inappropriate in a certain context. For example, a speaker may use the SOV order in (3b) because in that context, the speaker wants to *focus* the new object, Ahmet, and so places it in the immediately preverbal position. However, in (4)b, Ahmet is the *topic* or a link to the previous context whereas the subject, Fatma, is the *focus*, and thus the OSV word order is used. Here, we translate these Turkish sentences to English using different "stylistic" constructions (e.g. topicalization, it-clefts, phonological focusing etc.) in order to preserve approximately the same meanings.

(3) a. Fatma kimi arıyor?
    Fatma who seek-Pres?
    "Who is Fatma looking for?"
  b. Fatma Ahmet'i arıyor.      SOV
    Fatma Ahmet-Acc seek-Pres.
    "Fatma is looking for AHMET."

(4) a. Ahmet'i kim arıyor?
    Ahmet-Dat who seek-Pres.
    "Who is looking for Ahmet?"
  b.
    Ahmet'i Fatma arıyor.      OSV
    Ahmet-Acc Fatma seek-Pres.
    "As for Ahmet, it is FATMA who is looking for him."

It is very common for Turkish speakers to put information already mentioned in the discourse, i.e. discourse-given, in the post-verbal positions, in the *background* component of the information structure. In fact, discourse-new elements cannot occur in the postverbal positions. In addition, referential status, i.e. whether the speaker uses a full noun phrase, an overt pronoun, or a null pronoun to refer to an entity in the discourse, can be used to signal the familiarity information to the hearer. Thus, given information can be freely dropped (5)$b_1$ or placed in post-verbal positions (5)$b_2$ in Turkish. Although further research is required on the interaction between referential status and word order, I will not concentrate on this issue in this paper.

(5) a. Fatma Ahmet'i aradı.
    Fatma Ahmet-Acc seek-Past.
    "Fatma looked for Ahmet."
  $b_1$. Ama ∅ ∅ bulamadı.
    But ∅ ∅ find-Neg-Past.
    "But (she) could not find (him)."
  $b_2$. Ama bulamadı Fatma Ahmet'i.
    But find-Neg-Past Fatma Ahmet-Acc.
    "But she, Fatma, could not find him, Ahmet."

The same information structure components *topic, focus, background* can also explain the positioning of adjuncts in Turkish sentences. For example, placing a locative phrase in different positions in a sentence results in different discourse meanings, much as in English sentences:

(6) a. Fatma Ahmet'i Istanbul'da aradı.
    Fatma Ahmet-Acc Istanbul-loc seek-Past.
    "Fatma looked for Ahmet in ISTANBUL."
  b. Istanbul'da Fatma Ahmet'i aradı.
    Istanbul-loc Fatma Ahmet-Acc seek-Past.
    "In Istanbul, Fatma looked for Ahmet."
  c. Fatma Ahmet'i aradı Istanbul'da.
    Fatma Ahmet-Acc seek-Past Istanbul-loc.
    "Fatma looked for Ahmet, in Istanbul."

*Long distance scrambling*, word order permutation involving more than one clause, is also possible out of most embedded clauses in Turkish; in complex sentences, elements of the embedded clauses can occur in matrix clause positions. However, these word orders with long distance dependencies are only used by speakers for specific pragmatic functions. Generally, an element from the embedded clause can occur in the sentence initial topic position of the matrix clause, as in (7)b, or to the right of the matrix verb as backgrounded information, as in (7)c.[2]

(7) a.
    Fatma [Ayşe'nin gittiğini] biliyor.
    Fatma [Ayşe-Gen go-Ger-3sg-Acc] know-Prog.
    "Fatma knows that Ayşe left."
  b.
    Ayşe'nin$_i$ Fatma [$t_i$ gittiğini] biliyor.
    Ayşe-Gen$_i$ Fatma [$t_i$ go-Ger-3sg-Acc] know-Prog.
    "As for Ayşe, Fatma knows that she left."
  c.
    Fatma [$t_i$ gittiğini] biliyor Ayşe'nin$_i$.
    Fatma [$t_i$ go-Ger-3sg-Acc] know-Prog Ayşe-Gen$_i$.
    "Fatma knows that she, Ayşe, left."

## 3 The Categorial Formalism

Many different syntactic theories have been proposed to deal with free word order variation. It has been widely debated whether word order variation is the result of stylistic rules, the result of syntactic movement, or base-generated. I adopt a categorial framework in which the word order variations in Turkish are pragmatically-driven; this lexicalist framework is not compatible with transformational movement rules.

---
[2]I have put in coindexed traces and italicized the scrambled elements in these examples to help the reader; I am not making the syntactic claim that these traces actually exist.

My work is influenced by (Steedman-91) in which a theory of prosody, closely related to a theory of information structure, is integrated with Combinatory Categorial Grammars (CCGs). Often intonational phrase boundaries do not correspond to traditional phrase structure boundaries. However, by using the CCG type-raising and composition rules, CCG formalisms can produce nontraditional syntactic constituents which may match the intonational phrasing. These intonational phrases often correspond to a unit of planning or presentation with a single discourse function, much like the information structure components of topic, neutral, focus, and background in Turkish sentences. Thus, the ambiguity that CCG rules produce is not spurious, but in fact, necessary to capture prosodic and pragmatic phrasing. The surface structure of a sentence in CCGs can directly reflect its information structure, so that different derivations of the same sentence correspond to different information structures.

In the previous section, we saw that ordering of constituents in Turkish sentences is dependent on pragmatic functions, the information structure of the sentence, rather than on the argument structure of the sentence as in English. Moreover, information structure is distinct from argument structure in that adjuncts and elements from embedded clauses can serve a pragmatic function in the matrix sentence and thus be a component of the information structure without taking part in the argument structure of the matrix sentence. This suggests an approach where the ordering information which is dependent on the information structure is separated from the the argument structure of the sentence. In section 3.1, I describe a version of CCGs adapted for free word order languages in (Hoffman-92) to capture the argument structure of Turkish, while producing a flexible surface structure and word order. In addition, each CCG constituent is associated with a pragmatic counterpart, described in section 3.2, that contains the context-dependent word order restrictions.

## 3.1 {}-CCG

Multi-set Combinatory Categorial Grammars, {}-CCGs, (Hoffman-92) is a version of CCGs for free word order languages in which the subcategorization information associated with each verb does not specify the order of the arguments. Each verb is assigned a function category in the lexicon which specifies a *multi-set* of arguments, so that it can combine with its arguments in any order. For instance, a transitive verb has the following category $S|\{Nn, Na\}$ which defines a function looking for a set of arguments, nominative case noun phrase ($Nn$) and an accusative case noun phrase ($Na$), and resulting in the category $S$, a complete sentence, once it has found these arguments. Some phrase structure information is lost by representing a verb as a function with a set of arguments. However, this category is also associated with a semantic interpretation. For instance, the verb "see" could have the following category where the hierarchical information among the arguments is expressed within the semantic interpretation separated from the syntactic representation by a colon: $S : see(X, Y)|\{Nn : X, Na : Y\}$. This category can easily be transformed into a DAG representation like the following where coindices, $x$ and $y$, are indicated by italicized font.[3]

(8)
$$\left[\begin{array}{ll} \text{Result} & : \left[\begin{array}{ll} \text{Syn} & : [\text{Cat: S, Tense: Pres}] \\ \text{Sem} & : see(x,y) \end{array}\right] \\ \text{Args} & : \left\{\begin{array}{l} \left[\begin{array}{ll} \text{Syn} & : \left[\begin{array}{ll} \text{Cat} & : np \\ \text{Case} & : nom \end{array}\right] \\ \text{Sem} & : x \end{array}\right], \\ \left[\begin{array}{ll} \text{Syn} & : \left[\begin{array}{ll} \text{Cat} & : np \\ \text{Case} & : acc \end{array}\right] \\ \text{Sem} & : y \end{array}\right] \end{array}\right\} \end{array}\right]$$

We can modify the CCG application rules for functions with sets as follows. The sets indicated by braces in these rules are order-free, i.e. Y in the following rules can be any element in the set. Functions can specify a *direction* feature for each of their arguments, notated in the rules as an arrow above the argument.[4] We assume that a category $X|\{\ \}$ where $\{\ \}$ is the empty set rewrites by a clean-up rule to just $X$.

(9) a. **Forward Application' (>):**
    $X|\{\overrightarrow{Y},...\}\quad Y \Rightarrow X|\{...\}$
 b. **Backward Application' (<):**
    $Y \quad X|\{\overleftarrow{Y},...\} \Rightarrow X|\{...\}$

Using these new rules, a verb can apply to its arguments in any order. For example, the following is a derivation of a sentence with the word order Object-Subject-Verb[5]:

---
[3] To improve the efficiency of unification and parsing, the arguments in the set can be associated with feature labels which indicate their category and case.

[4] Since Turkish is not strictly verb-final, most verbs will not specify the direction features of their arguments.

[5] Since I adopt a bottom-up generation algorithm, these derivations are used in both the parsing and the generation of Turkish sentences.

(10) Gazeteyi      Ayşe      okuyor.
    Newspaper-acc  Ayşe      reads.
    Na             Nn        S|{Nn,Na}
    ─────────────────────────────── <
                   S|{Na}
    ─────────────────────────────── <
                   S

Instead of using the set notation, we could imagine assigning Turkish verbs multiple lexical entries, one for each possible word order permutation; for example, a transitive verb could be assigned the categories $S\backslash Nn\backslash Na$, $S\backslash Na\backslash Nn$, $S\backslash Na/Nn$, etc., instead of the one entry $S|\{Nn, Na\}$. However, we will see below that the set notation is more than a shorthand representing multiple entries because it allows us to handle long distance scrambling, permutations involving more than one clause, as well.

The following composition rules are proposed to combine two functions with set-valued arguments, e.g. two verbs.

(11) a. **Forward Composition' ($> B$):**
        $X|\{\overrightarrow{Y}, ..._1\}$   $Y|\{..._2\}$   $\Rightarrow$   $X|\{..._1, ..._2\}$
     b. **Backward Composition' ($< B$):**
        $Y|\{..._1\}$   $X|\{\overleftarrow{Y}, ..._2\}$   $\Rightarrow$   $X|\{..._1, ..._2\}$

These composition rules allow two verb categories with sets of arguments to combine together. For example,

(12)
    go-gerund-acc                  knows.
    $S_{Na} : go(y)|\{Ng : y\}$    $S : know(x,p) |\{Nn: x, S_{na}: p\}$
    ──────────────────────────────────────────────────── <B
              $S : know(x, go(y))|\{Ng : y, Nn : x\}$

As the two verbs combine, their arguments collapse into one argument set in the syntactic representation. However, the verbs' respective arguments are still distinct within the semantic representation of the sentence. The predicate-argument structure of the subordinate clause is embedded into the semantic representation of the matrix clause. Long distance scrambling can easily be handled by first composing the verbs together to form a complex verbal function which can then apply to all of the arguments in any order.

Certain coordination constructions (such as 'SO and SOV' seen in (13) as well as 'SOV and SO') can be handled in a CCG based formalism by type-raising NPs into functions over verbs. Two type-raised noun phrases can combine together using the composition rules to form a nontraditional constituent which can then coordinate.

(13)    Ayşe kitabı,     Fatma da  gazeteyi      okuyor.
        Ayşe book-acc,   Fatma too newspaper-acc reads.
        "Ayşe is reading the book and Fatma the newspaper."

Order-preserving type-raising rules that are modified for {}-CCGs are used to convert nouns in the grammar into functors over the verbs. These rules can be obligatorily activated in the lexicon when case-marking morphemes attach to the noun stems.

(14) a. N + case $\Rightarrow (S|\{...\}) | \{S|\{\overleftarrow{Ncase}, ...\}\}$ $>$
     b. N + case $\Rightarrow (S|\{...\}) | \{S |\{\overrightarrow{Ncase}, ...\}\}$ $<$

The first rule indicates that a noun in the presence of a case morpheme becomes a functor looking for a verb on its right; this verb is also a functor looking for the original noun with the appropriate case on its left. After the noun functor combines with the appropriate verb, the result is a functor which is looking for the remaining arguments of the verb. The notation ... is a variable which can unify with one or more elements of a set.

The second type-raising rule indicates that a case-marked noun is looking for a verb on its left. {}-CCGs can model a strictly verb-final language like Korean by restricting the noun phrases of that language to the first type-raising rule. Since most, but not all, case-marked nouns in Turkish can occur behind the verb, certain pragmatic and semantic properties of a Turkish noun determine whether it can type-raise to the category produced by the second rule.

The {}-CCG for Turkish described above can be used to parse and generate all word orders in Turkish sentences. However, it does not capture the more interesting questions about word order variation: namely, *why* speakers choose a certain word order in a certain context and what additional meaning these different word orders provide to the hearer. Thus, we need to integrate the {}-CCG formalism with a level of information structure that represents pragmatic functions, such as topic and focus, of constituents in the sentence in a compositional way.

### 3.2 A Grammar for Word Order

In (Steedman-91; Prevost/Steedman-93), a theory of prosody, closely related to a theory of information structure, is integrated with CCGs by associating every CCG category encoding syntactic and semantic properties with a prosodic category. Taking advantage of the non-traditional constituents that CCGs can produce, two CCG constituents are allowed to combine only if their prosodic counterparts can also combine.

Similarly, I adopt a simple interface between {}-CCG and ordering information by associating each syntactic/semantic category with an ordering category which bears linear precedence information. These two categories are linked together by the features of the information structure. For example, the verb "arıyor"

$$\text{arıyor (seek)} := \begin{bmatrix} \text{Category} : \begin{bmatrix} \text{Result} : \begin{bmatrix} \text{Syn} : \begin{bmatrix} \text{Cat} & : & S \\ \text{Tense} & : & \text{Pres} \end{bmatrix} \\ \text{Sem} : \begin{bmatrix} \text{Event} & : & e \\ \text{LF} & : & [\text{seek}(x,y), Xlf, Ylf] \end{bmatrix} \\ \text{Info} : I \begin{bmatrix} \text{Theme} & : & \begin{bmatrix} \text{Topic} & : & T \\ \text{Neutral} & : & N \end{bmatrix} \\ \text{Rheme} & : & \begin{bmatrix} \text{Focus} & : & F \\ \text{Main-Prop} & : & seek \end{bmatrix} \\ \text{Background} & : & B \end{bmatrix} \end{bmatrix} \\ \text{Args} : \left\{ \begin{bmatrix} \text{Syn} : \begin{bmatrix} \text{Cat} & : & \text{np} \\ \text{Case} & : & \text{nom} \end{bmatrix} \\ \text{Sem} : \begin{bmatrix} \text{Entity} & : & x \\ \text{Props} & : & Xlf \end{bmatrix} \end{bmatrix}, \begin{bmatrix} \text{Syn} : \begin{bmatrix} \text{Cat} & : & \text{np} \\ \text{Case} & : & \text{acc} \end{bmatrix} \\ \text{Sem} : \begin{bmatrix} \text{Entity} & : & y \\ \text{Props} & : & Ylf \end{bmatrix} \end{bmatrix} \right\} \end{bmatrix} \\ \text{Order} : \quad I/([\text{Background} : B])\backslash([\text{Topic} : T])\backslash([\text{Neutral} : N])\backslash([\text{Focus} : F]) \end{bmatrix}$$

Figure 1: The Lexical Entry for a Transitive Verb, "arıyor" (seeks).

(seeks) is assigned the lexical entry seen in the *category* feature of the DAG in Figure 1. The *category* feature contains the argument structure in the features *syn* and *sem* as well as the information structure in the feature *info*. This lexical entry is associated with an ordering category seen in the feature *order* of the DAG in Figure 1. This ordering feature is linked to the *category* feature via the co-indices $T$, $N$, $F$, and $B$.

The ordering categories are assigned to lexical entries according to context-dependent word order restrictions found in the language. All Turkish verbs are assigned the ordering category seen in the *order* feature in Figure 1; this is a function which can use the categorial application rules to first combine with a focused constituent on its left, then a neutral constituent on its left, then a topic constituent on its left, and then a background constituent on its right, finally resulting in a complete utterance. This function represents the template mentioned in example 2 for assigning discourse functions according to their positional relation to the verb following (Erguvanli-84). However, it is more flexible than Erguvanlı's approach in that it allows more than one possible information structure. The parentheses around the arguments of the ordering category indicate that they are optional arguments. The sentence may contain all or some or none of these information structure components.[6] It may turn out that we need to restrict the optionality on these components. For instance, if there is no topic found in the sentence-initial position, then we may need to infer a topic or a link to the previous context. In the current implementation, the focus is an obligatory constituent in order to ensure that the parser produces the derivation with the most likely information structure first, and there is an additional ordering category possible where the verb itself is focused and where there are no pre-verbal elements in the sentence.

Categories other than verbs, such as nouns, determiners, adjectives, and adverbs, are associated with an ordering category that is just a basic element, not a function. In Turkish, the familiarity status of entities in the discourse model serves a role in determining their discourse function. For example, discourse-new entities cannot occur in the post-verbal or sentence initial positions in Turkish sentences. Thus, discourse-new elements can be assigned ordering categories with the feature-attribute *focus* or *neutral* with their semantic properties as the feature-value, but they cannot be associated with *background* or *topic* ordering categories. There are no such restrictions for discourse-old entities; thus they can be assigned a variable which can unify with any of the information structure components.

During a derivation in parsing or generation, two constituents can combine only if the categories in their *category* features can combine using the {}-CCG rules presented in the previous section, *and* the categories in their *order* features can combine using the following rewriting rules. A sample derivation involving the ordering grammar can be seen in Figure 2.

---

[6] This is similar to Vallduvi's approach (Vallduvi-90) for Catalan in which there are three components, *link* (to the previous context), *focus*, and *tail*, where the link and tail can be optional. One difference is that Vallduvi allows the verb to be a part of any component, but it is not clear how this could be implemented.

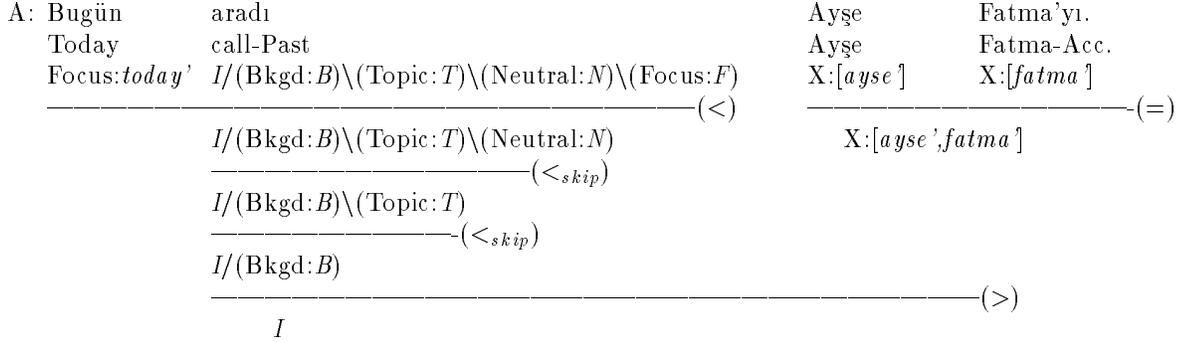

Figure 2: A Derivation involving just the Ordering Categories.

(15) a. **Forward Application ($>$):**
$X/Y \quad Y \Rightarrow X$ where Y is not a functor.
b. **Backward Application ($<$):**
$Y \quad X\backslash Y \Rightarrow X$ where Y is not a functor.
c. **Forward Skip-Optional Rule ($>_{skip}$):**
$X/(Y) \quad Z \Rightarrow X \ Z$
d. **Backward Skip-Optional Rule ($<_{skip}$):**
$Z \quad X\backslash(Y) \Rightarrow X \ Z$
e. **Identity ($=$):**
$X \quad X \Rightarrow X$

The identity rule allows two constituents with the same discourse function to combine. The resulting constituent may not be a traditional syntactic constituent, however as argued in (Steedman-91), this is where we see the advantage of using a CCG based formalism. Through type-raising and composition, CCG formalisms can produce nontraditional syntactic constituents which may have a single discourse function. For example in Figure 2, the NPs *Fatma* and *Ayşe* form a pragmatic constituent using the identity rule in the ordering grammar; in order to form a syntactic constituent as well, they must be type-raised and combine together using the {}-CCG composition rule. Type-raising in Turkish is needed for sentences involving more than one NP in the *neutral* and *background* positions.

The ordering grammar also allows adjuncts and elements from other clauses (long distance scrambled) to be components in the information structure. This is because the information structure in a verb's lexical entry does not specify that its components must be arguments of the verb in its argument structure. Thus, adjuncts and elements from embedded clauses can be serve a purpose in the information structure of the matrix clause, although they are not subcategorized arguments of the matrix verb. For long distance scrambling, the additional restriction (that Y is not a functor) on the application rules given above ensures that a verb in the embedded clause has already combined with all of its obligatory arguments or skipped all of its optional arguments before combining with the matrix verb.

The ordering grammar presented above is similar to the template grammars in (Danlos-87), the syntax specialists in PAULINE (Hovy-88), and the realization classes in MUMBLE (McDonald/Pustejovsky-85) in that it allows certain pragmatic distinctions to influence the syntactic construction of the sentence. The ordering grammar does not make as fine-grained pragmatic distinctions as the generation systems above, but it represents language-specific and context-dependent word order restrictions that can be lexicalized into compositional categories. The categorial formalism presented above captures the general discourse meaning of word order variation in languages such as Turkish while using a compositional method.

## 4 The Implementation

I have implemented a simple data-base query task, diagramed in Figure 3, to demonstrate how the categorial formalism presented in the previous section can generate Turkish sentences with word orders appropriate to the context. The system simulates a Personal Assistant who schedules meetings and phone calls with a number of individuals. The user issues queries to which the program responds, after consulting the data-base, in sentences with the appropriate word order, while maintaining a model of the changing context.

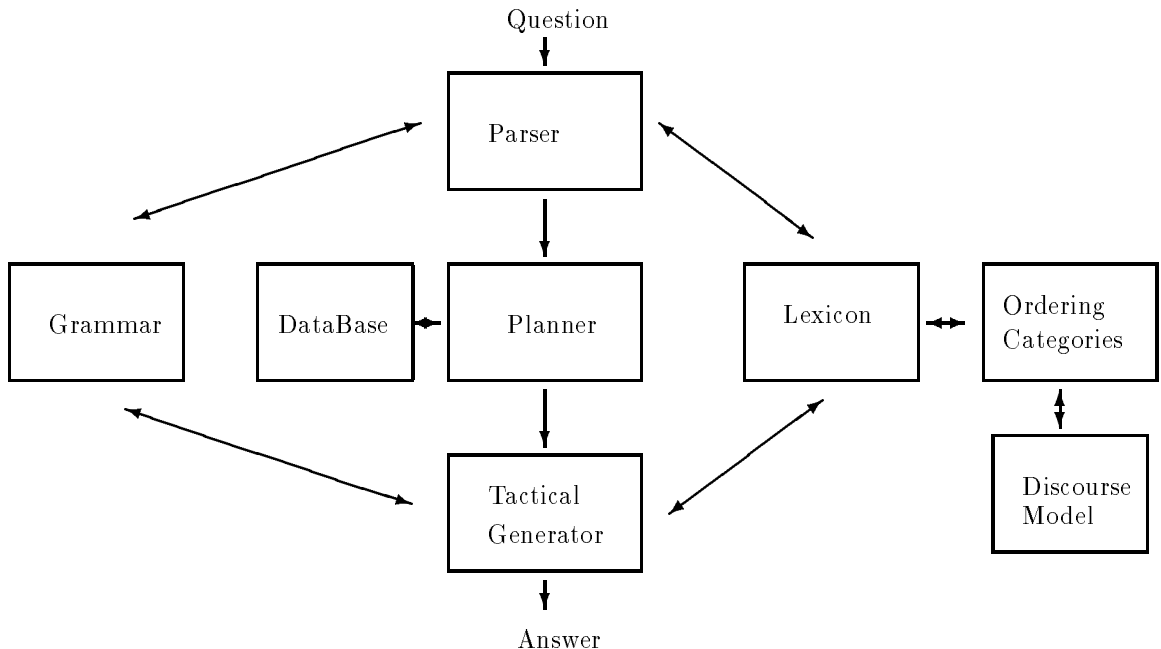

Figure 3: The Personal Assistant Generation System

Since most of the information is lexicalized, the same grammar and lexicon is used by the parser and the generator. After the question is parsed, the discourse model is updated[7], and the question's representation is sent to the planning component of the generator. The planner at this point consists of simple plans for constructing answers to certain wh-questions and yes/no questions. Certain predicates in the queries trigger the planner to look up schedules and make appointments for the agents mentioned in the query.

The planner creates a representation for the answer by copying much of the question representation and by adding the appropriate new information found in the database. The information structure of the question can be used by the planner as well. The topic of the question is copied to the answer in order to maintain topic continuity, although in a less limited domain, a separate algorithm is needed to allow for shifts in topic. In addition, when a yes/no question is not validated in the data-base, the planner replaces the focus of the question with a variable and requests another search of the data-base to find a new focus which statisfies the rest of the question. For example,[8]

---

[7] As suggested by (Vallduvi-90), the information structure of a sentence can provide cues on how to update and organize the discourse model.

[8] Particles such as "yes" and "no" are are produced by a separate call to the generator, before generating the answer.

(16) a. Ahmet Fatma'yi gördü mü?
     Ahmet Fatma-Acc see-Past Quest?
     "Did Ahmet see FATMA?"
  b. Hayır, ama Ahmet Ayşe'yi gördü.
     No, but Ahmet Ayşe-Acc see-Past.
     "No, but Ahmet saw AYŞE.

In all question types, the information found in the database lookup is specified to be the *focus* of the answer. The semantic properties of the focused entity are either found in the database, or if it has already been mentioned in the discourse, by consulting the discourse model. The planner then passes the representation of the answer to the realization component of the generator described in the next section.

## 4.1 Head-driven Bottom-up Generation

I adopt a head-driven bottom up generation algorithm (Calder/etal, 1989; Shieber/etal-89; vanNoord-90) that takes advantage of lexical information as well as the top-down input provided by the planner. This approach is particularly useful for categorial grammars since most of the information is stored in the lexical entries rather than the grammar rules.

The planning component described above provides the input for the algorithm, for example, to generate a sentence with the syntactic, semantic, and information structure features shown in Figure 4.[9] The input does not have to fully specify the word order in the information structure. For instance, since the description in Figure 4 of the sentence to be generated does not specify the function of "the students" in the information structure, either of the following two word orders can be generated:

(17) a. Ayşe öğrencileri üçte gördü.
    Ayşe student-Pl-Acc three-Loc see-Past.
    "Ayşe saw the students at THREE."
  b. Ayşe üçte gördü öğrencileri.
    Ayşe three-Loc see-Past student-Pl-Acc.
    "Ayşe saw the students at THREE."

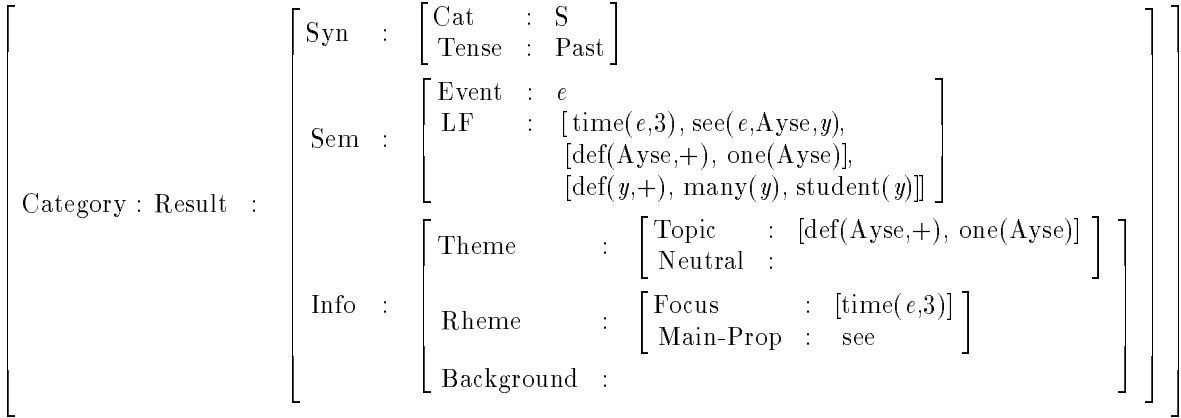

Figure 4: Input to the Generation Algorithm.

The algorithm for the head-driven bottom up generator is seen below:

```
generate(Input) :-
      find_lex_cat(Input,LexDag),
      bup_generate(Input,LexDag).

bup_generate(Input,LexDag):- unify(Input,LexDag).

bup_generate(Input, LexDag) :-
        combine(Arg, LexDag, ResDag, backward),
        generate(Arg),
        order(Arg, LexDag, ResDag),
        concat_phons(Arg, LexDag, ResDag),
        bup_generate(Input, ResDag).

bup_generate(Input, LexDag) :-
        combine(LexDag, Arg, ResDag, forward),
        generate(Arg),
        order(LexDag, Arg, ResDag),
        concat_phons(LexDag, Arg, ResDag),
        bup_generate(Input, ResDag).
```

This algorithm is very similar to the (Calder/etal, 1989) algorithm for Unificational Categorial Grammar (UCG). First, the function **generate** finds a category in the lexicon which is the head of the sentence. Then in **bup-generate**, we try to apply the combinatory grammar rules (i.e. the forward and backward {}-CCG rules) to this lexical functor to generate its arguments in a bottom-up fashion. The order function applies the ordering rules to the functor and argument to make sure that they form a constituent in the information structure. The **bup-generate** function is called recursively on the result of applying the rules until it has found all of the head functor's ($LexDag$) arguments, eventually resulting in something which unifies with the $Input$.[10]

The main difference between this CCG algorithm and the UCG algorithm is that the CCG algorithm uses all of the information (syntactic, semantic, and information structure features) given in the input, instead of using only the semantic information, to find the head functor in the lexicon. This is possible because of the formulation of the CCG rules. We can assume there is some function in the lexicon whose result unifies with the input, if this function is to take part in a CCG derivation that produces the input. This assumption is built into the CCG rules, since the head daughter in each rule (shown in bold in the following {}-CCG rules) shares its function result (X) with the final result after applying the rule:

(18) a. $\boldsymbol{X|\{\overrightarrow{Y},...\}}$  $Y$ $\Rightarrow X|\{...\}$
   b. $Y$  $\boldsymbol{X|\{\overleftarrow{Y},...\}}$ $\Rightarrow X|\{...\}$

---

[9]Note that the semantic predicates of the sentence are represented using a list notation; the DAG unification algorithm has been extended to recognize the function format such as $student(x)$ as features.

[10]Note that order and concat-phons must be called after we have lexically instantiated both Arg and LexDag to avoid infinite loops. The UCG algorithm also freezes such features until the argument is instantiated.

c. $X|\{\overrightarrow{Y}, ..._1\}$ $Y|\{..._2\} \Rightarrow X|\{..._1, ..._2\}$

d. $Y|\{..._1\}$ $X|\{\overleftarrow{Y}, ..._2\} \Rightarrow X|\{..._1, ..._2\}$

To make the algorithm more efficient, **find-lex-cat** first finds a rough match in the lexicon using term-unification. We associate each item in the lexicon with a semantic key-predicate that is one of the properties in its semantic description. A lexical entry roughly matches the input if its semantic key-predicate is a member of the list of semantic properties given in the input. After a rough match using term-unification, **find-lex-cat** unifies the DAGs containing all of the known syntactic, semantic, and pragmatic information for the most embedded result of the lexical category and the result of the input, e.g. Figure 4, to find the lexical category which is the head functor.[11] Then, the rules can be applied in a bottom up fashion assuming that the found lexical category is the head daughter in the rules.

In this section, I have shown how the head-driven bottom-up generation algorithm can be adapted for the CCG formalism. The following sample runs of the generation system further demonstrate how context-appropriate word orders are generated in this formalism.

## 4.2 Sample Runs

The sample runs below represent the following translated dialogue:

(19) a. Fatma Ayşe'yi görebilirmi?
Fatma Ayşe-Acc see-abil-aor-quest?
"Can Fatma see Ayşe?"

b. Evet, Fatma Ayşe'yi ikide görebilir.
Yes, Fatma Ayşe-Acc two-Loc see-abil-aor.
"Yes, Fatma can see Ayşe at TWO."

(20) a. Ikide kimi görecek Fatma?
Two-Loc who-Acc see-Future Fatma?
"Who will Fatma see at two?"

b. Ikide Ayşe'yi görecek Fatma.
Two-loc Ayşe-Acc see-Pres-prog Fatma.
"At two, she, Fatma, will see AYŞE."

The questions are the user's input and the answers are the Personal Assistant system's replies. Each question-answer pair is followed by the DAG representation of the answer. Note that the syntax and semantics of the answers in each run are identical, but the difference in their information structure is reflected in the word order that is generated.

---

[11] The function find-lex-cat can also find type-raised NPs in the lexicon whose results unify with the input. To prevent infinite loops, I represent the semantic features of the NP in the result of the type-raised function, rather than having a variable $v$ as the result in $v/(v\backslash np)$.

```
|: fatma ayseyi gorebilirmi?
Answer: evet, fatma ayseyi ikide gorebilir.

 Dag:
 syn :
    cat : s
    voice : active
    tense : aorist
    agr :
         number : sing
         person : 3
    compound : abilitive
 sem :
    type : decl
    lf : [time(e6,2), see(e6,fatma,ayse),
          [one(fatma),def(fatma,+)],
          [one(ayse),def(ayse,+)]]
    event : e6
 info :
    rheme :
           focus : [time(e6,2)]
           mainprop : see(e6,fatma,ayse)
    theme :
           topic : [one(fatma), def(fatma,+)]
           neutral : [one(ayse), def(ayse,+)]
    background : none

|: ikide  kimi  gorecek  fatma?
Answer: ikide ayseyi gorecek fatma.
 Dag:
 syn :
    cat : s
    voice : active
    tense : future
    agr :
         number : sing
         person : 3
 sem :
    type : decl
    lf : [time(e6,2), see(e6,fatma,ayse),
          [one(fatma),def(fatma,+)],
          [one(ayse),def(ayse,+)]]
    event : e6
 info :
    rheme :
           focus :  [one(ayse),def(ayse,+)]
           mainprop : see(e6,fatma,ayse)
    theme :
           topic : [time(e6,2)]
           neutral : none
    background : [one(fatma),def(fatma,+)]
```

# 5 Conclusions

In this paper, I have presented a strategy for the realization component of a generation system to handle word order variation in Turkish sentences. I integrated a level of information structure with a unification-based version of Combinatory Categorial Grammars, adapted for free word order languages. I discussed an implementation of a database query task using a modified head-driven bottom-up generation algorithm to demonstrate how the categorial formalism generates Turkish sentences with word orders appropriate to the context.

Further research is needed on processing the information found in the information structure after parsing a sentence, e.g. inferences about how focused discourse entities or topic entities are related to sets of other discourse entities in the discourse model. In addition, a separate algorithm, perhaps Centering Theory (Grosz/etal-83), is needed to keep track of the salience of discourse entities and resolve the reference of empty pronouns, or in the case of generation, to determine what must be realized and what can be dropped in the utterance. In future research, I would also like to extend this same approach to generate certain stylistic constructions in English such as topicalization, it-clefts, and right dislocation.